\documentclass[11pt,twocolumn,letterpaper]{article}

\usepackage{cvpr}

\usepackage{amsmath}
\usepackage{amssymb}
\usepackage{booktabs}
\usepackage{caption}
\usepackage{epsfig}
\usepackage{graphicx}
\usepackage{overpic}
\usepackage{subfigure}
\usepackage{times}

\usepackage{enumitem}
\setenumerate[1]{itemsep=0pt,partopsep=0pt,parsep=\parskip,topsep=5pt}
\setitemize[1]{itemsep=0pt,partopsep=0pt,parsep=\parskip,topsep=5pt}
\setdescription{itemsep=0pt,partopsep=0pt,parsep=\parskip,topsep=5pt}

\usepackage[pagebackref=true,breaklinks=true,letterpaper=true,colorlinks,bookmarks=false]{hyperref}

\cvprfinalcopy % *** Uncomment this line for the final submission

\def\cvprPaperID{159} % *** Enter the CVPR Paper ID here

\newcommand{\mypara}[1]{\paragraph{#1.}}

\graphicspath{{figures/}}

\begin{document}

%%%%%%%%% TITLE

\title{Influenza Modeling Based on Massive Feature Engineering \\ and International Flow Deconvolution~\footnotemark[1]}

\author{Ziming Liu \quad Yixuan Wang \quad Zizhao Han \quad Dian Wu}

\author{%
  Ziming Liu$^{\mathsection}$\footnotemark[2]\quad Yixuan Wang$^\ddagger$\footnotemark[2]\quad Zizhao Han$^{\mathsection}$\footnotemark[2]\quad and Dian Wu$^{\mathsection}$\footnotemark[2]\\
    $^\mathsection$School of Physics, Peking University, Beijing, China \\
  $^\ddagger$School of Mathematical Sciences, Peking University, Beijing, China\\
  E-mails: \texttt{liu\_zi\_ming@pku.edu.cn, roywangyx@pku.edu.cn,}\\ \texttt{zzhan@pku.edu.cn, wdphy16@pku.edu.cn} 
 }

\maketitle
\renewcommand{\thefootnote}{\fnsymbol{footnote}}
\footnotetext[1]{This paper serves as a submitted work to the Citadel Data Open Final Competition at The New York Stock Exchange (NYSE) in April, 2019. We participated in this international event as the champion of the regional competition in Beijing.}
\footnotetext[2]{These authors contributed equally to this work.}

%%%%%%%%% ABSTRACT
\begin{abstract}
In this article, we focus on the analysis of the potential factors driving the spread of influenza, and possible policies to mitigate the adverse effects of the disease. To be precise, we first invoke discrete Fourier transform (DFT) to conclude a yearly periodic regional structure in the influenza activity, thus safely restricting ourselves to the analysis of the yearly influenza behavior. Then we collect a massive number of possible region-wise indicators contributing to the influenza mortality, such as consumption, immunization, sanitation, water quality, and other indicators from external data, with $1170$ dimensions in total. We extract significant features from the high dimensional indicators using a combination of data analysis techniques, including matrix completion, support vector machines (SVM), autoencoders, and principal component analysis (PCA). Furthermore, we model the international flow of migration and trade as a convolution on regional influenza activity, and solve the deconvolution problem as higher-order perturbations to the linear regression, thus separating regional and international factors related to the influenza mortality. Finally, both the original model and the perturbed model are tested on regional examples, as validations of our models.
% TODO: rewrite below
Pertaining to the policy, we make a proposal based on the connectivity data along with the previously extracted significant features to alleviate the impact of influenza, as well as efficiently propagate and carry out the policies. We conclude that environmental features and economic features are of significance to the influenza mortality. The model can be easily adapted to model other types of infectious diseases.
\end{abstract}

%%%%%%%%% BODY TEXT %%%%%%%%%%%%%%%%%%%%%%%%%%%%%%%%%%%%%%%%
\section{Introduction}\label{sec:Introduction}

Infectious diseases pose an incessant threat to human health and welfare. Influenza, as one of the most prevalent diseases worldwide, is a typical recurrent seasonal epidemic disease. We shall seek to model the pattern of its behavior and its causes, based on the provided data and external data online. Qualitative patterns are proposed, supplemented a quantitative feature extraction along with its higher-order rectified version, to analyze the possible causes of influenza. Concrete policies are provided, both regarding its effectiveness and propagation viability, after a detailed regional validation of our models.

%%%%%%%%%%%%%%%%%%%%%%%%%%%%%%%%%%%%%%%%%%%%%%%%%%%%%%%%%%%%%%%%%%%%%%%%%%%%%%%%%
\section{General Idea and Analysis}\label{sec:General Idea}

\begin{figure*}[tb]
\centering\includegraphics[width=\linewidth]{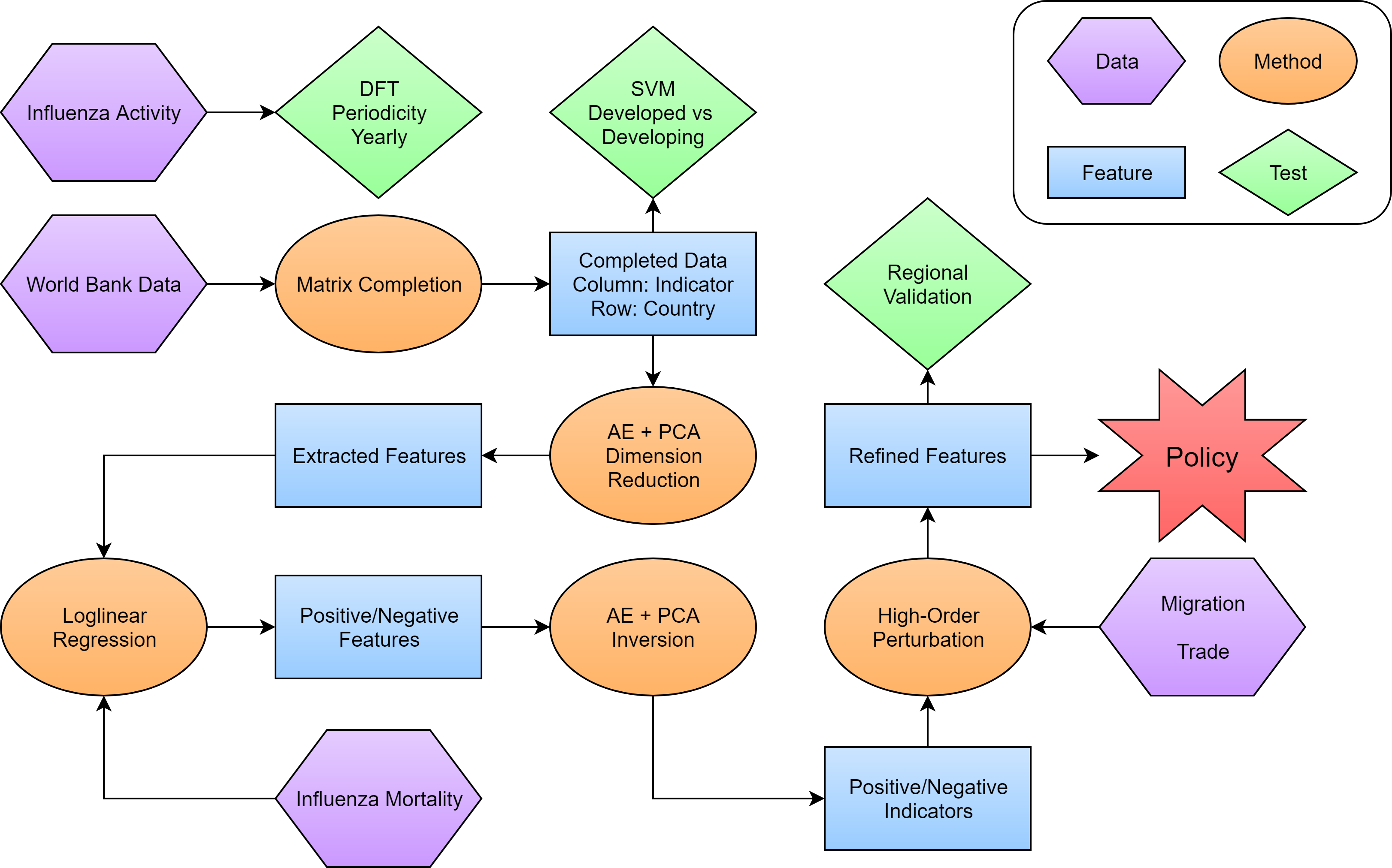}
\caption{Pipeline for our model}
\label{fig:pipeline}
\end{figure*}

The organization of our model is mainly composed of five parts. First, we need to preprocess the data and carefully select the relevant information. Then we shall analyze the property of the available data qualitatively and build our model as per the indication and real-life scenarios. Thirdly, the principal features shall then be extracted and their validity shall be tested against a priori criteria. Now we can use such features to fit our models and derive the weights of each feature. A higher-order perturbed version of the model is also introduced to address the international behavior, and we shall test the effectiveness of both models. Finally, we design the corresponding policy to prevent the spread of influenza as suggested by our main model. The pipeline for our models is plotted in Figure~\ref{fig:pipeline}.

\mypara{Data Augmentation} Based on our formulation of the problem, we first augment the data related to the influenza mortality. Besides the provided data on consumption, health indicators, connectivity, immunization, sanitation, and water quality, we also resort to the World Bank Open Data~\cite{xixi} to collect other possible indicators, such as labor or agriculture. In total, we make a massive collection of indicators with $1170$ dimensions. The rationale behind this resort to external data is due to the sparsity and insufficiency of our provided data. Although there are still missing values in the collected indicators, we can obtain a dense dataset via matrix completion~\cite{mazumder2010spectral}. Finally, we shall use support vector machines (SVM)~\cite{suykens1999least} to certify that our augmented dataset is indeed consistent and satisfies the basic criteria.

\mypara{Qualitative Analysis of Periodic Behavior} Since influenza, like all recurrent infectious diseases, follows a somewhat periodic behavior, we have sound reason to conjecture a periodic outbreak of influenza. Discrete Fourier transforms~\cite{weinstein1971data} on weekly influenza activity both regionally and globally would reveal such pattern indeed, and we find the period to be approximately one year. Thus we can safely use only the provided annual data, as it is the representative of the mean of influenza activity each year.

\mypara{Feature Extraction} We shall use a combination of autoencoders~\cite{ng2011sparse} and principal component analysis (PCA)~\cite{abdi2010principal} to extract useful features from the augmented data. Autoencoders are used to introduce nonlinearity and thus robustness to our model. Then PCA is introduced for an orthogonal design. We shall use autoencoders to extract our features to a $10$ dimensional Riemann manifold, and use PCA to analyze its $6$ principal characteristics. Our method is in effect a kernel PCA method which enjoys both orthogonality and nonlinearity.

\mypara{Regional Model} To analyze the causes of influenza based on extracted features, we propose a simple linear regression to fit the weights based on the influenza mortality data, as opposed to prevalent models~\cite{webster1992evolution} of PDE / integro-differential equations. The rationale behind the choice of the model is analyzed as follows. On the one hand, we do not possess a sufficiently fine resolution of data points in time at such a global scale, thus rendering a continuous model errant. On the other hand, nonlinearity is already introduced in our data pre-processing and feature extracting process. Therefore a stable linear model would appear more rational and can be easily modified by international higher-order perturbation terms.

\mypara{Global Model} Taken into account the global migration of population and trade of products, we can propose a rectified version of the regional model using the provided migration data, and trade data from Trade Map~\cite{qaq}. Infectious diseases can indeed spread via migration and trade, so we model the international flow of migration and trade as a convolution on regional influenza activity. Furthermore, we use a second-order polynomial to model the region-wise dependence of the convolution kernel. We jointly optimize the coefficients of the deconvolution and the linear regression towards the correct prediction. Regional data shall be used to verify the correctness of our models.

\mypara{Policy Design} Based on the analysis of features and their importance, we shall design corresponding policies to mitigate the negative influences of influenza. Cost-effectiveness and viability shall also be taken into consideration when concrete policies are designed for each region.

\section{Main Models and Guiding Methods}

\subsection{Discrete Fourier Transform}

Fourier analysis is used to analyze the behavior of a continuous function in the frequency domain~\cite{grafakos2008classical}. Specifically, Fourier transform is a linear bijective mapping of the $L^2$ space to itself. When given discrete data in a finite interval, Discrete Fourier Transform~\cite{harris1978use} as a variant of such methods can be used to analyze the periodic behavior quantitatively.

For any given region, we consider the available data on influenza activity. For the known $N$ data points of weekly influenza activity $x_1, x_2, \cdots, x_N$, we use the DFT technique to get coefficients:
\begin{equation}
X[k] = \sum_{n = 1}^{N} x_n \mathrm{e}^{-\mathrm{i} 2 \pi k n / N} \quad \text{for} \, 1 \leq k \leq N.
\end{equation}
We can infer the periodic behavior of influenza activity based on the peaks and periodic decaying property of the coefficients $X[k]$.

\subsection{Matrix Completion}

Matrix completion~\cite{mazumder2010spectral} is a conventional technique for filling out missing information in a matrix of a large scale, with the aim of maintaining a sparse and low-rank structure. The precise formulation of the problem is defined as follows:
\begin{equation}
\begin{array}{ll}
{\min_X} & {\operatorname{rank}(X)} \\
{\text{subject to}} & {X_{i j} = M_{i j} \quad \forall M_{i j} \in \text{known} \, M}.
\end{array}
\end{equation}
Though as an NP-hard problem to find the optimal completion, we can introduce regularization and relaxation to address the problem, even with noisy input. We shall resort to the Python package $\texttt{fancyimpute}$ for a numerical implementation of completing the missing data from collected indicators relevant to the influenza mortality.

\subsection{Support Vector Machine}

Given $N$ data points $\left\{ y_i, x_i \right\}_{i = 1}^N$, where $x_i \in \mathbb{R}^N$ is the input feature and $y_i$ is the output pattern, assumed to be $1$ or $-1$, support vector method approach~\cite{suykens1999least} aims at a construction of the classifier in the following form:
\begin{equation}
w^T x - b = 0.
\end{equation}
Namely, we seek to separate the data labels by a separating hyperplane. The optimal parameters are achieved for hyperplanes minimizing the hinge loss function
\begin{equation}
\left[ \frac{1}{N} \sum_{i = 1}^N \max \left(0, 1 - y_i \left( w^T x_i - b \right) \right) \right] + \lambda \|w\|^2,
\end{equation}
where $\lambda$ is a tuning regularization parameter. We shall use SVM and a priori knowledge to verify the results after matrix completion.

\subsection{Autoencoder}

\begin{figure*}[tb]
\centering\includegraphics[width=\linewidth]{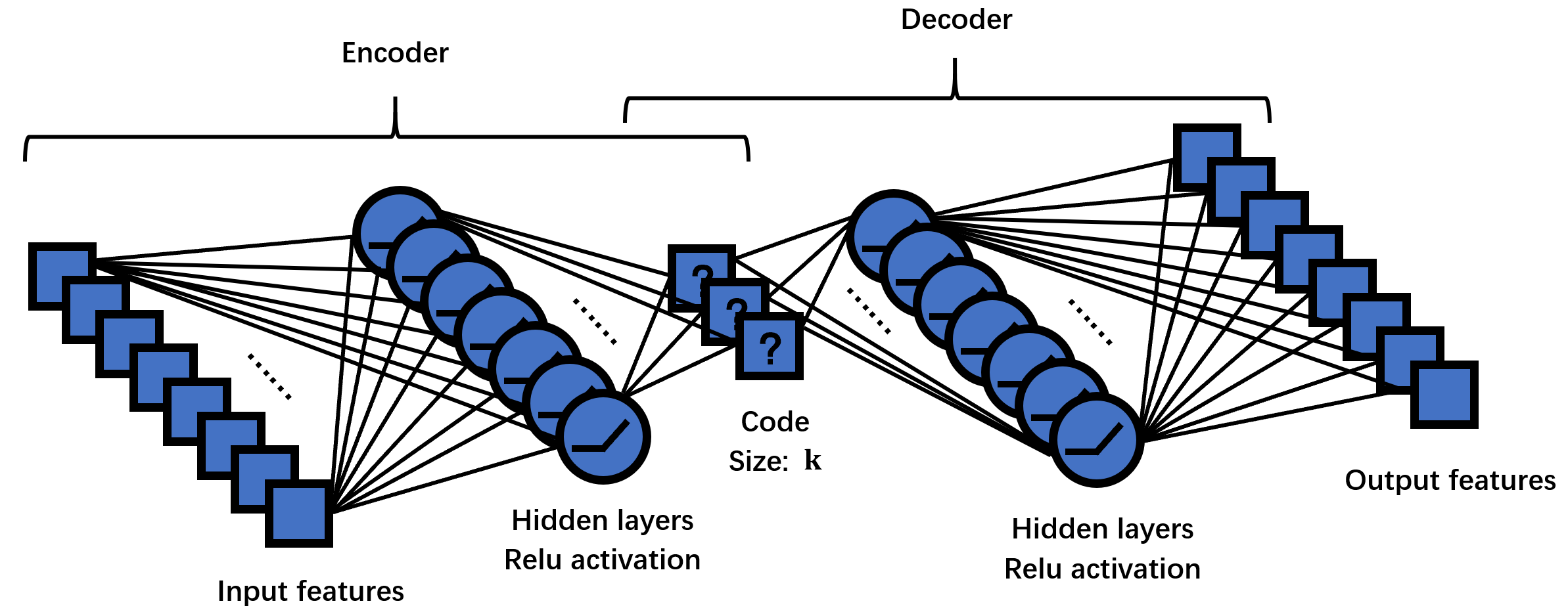}
\caption{Architecture of our autoencoder}
\label{fig:auto}
\end{figure*}

Recent years have witnessed the success of variants of autoencoders~\cite{ng2011sparse}, mainly because autoencoders have the power to express and represent nonlinear manifold in a high dimensional space. We use autoencoders to reduce the dimension of collected indicators, which can be modeled by the `bottleneck' structure. To be specific, the architecture is plotted in Figure~\ref{fig:auto}. The encoder part have $1170$ cells as inputs, and multiple hidden full connection layers, with ReLU activation function to introduce nonlinearity. A `bottleneck' layer with $k$ cells ($k \ll 1170$) is reached in the middle. The decoder part is a mirror of the encoder, with output size $1170$. The features in the `bottleneck' layer are recognized as the `code' which most likely represent the most important information hidden in data, thus the name `autoencoder'.

In order to choose the optimal number of layers and number of cells in the `bottleneck' layer, therefore balance between the reconstruction error and the model simplicity, we use the Bayesian information criterion (BIC)~\cite{chen2008extended} to evaluate the overall performance of our model. Here, the BIC factor is defined as
\begin{equation}
\mathrm{BIC} = \ln(n) k + 2 n L,
\end{equation}
where $n = 183$ is the number of countries and regions in context, $k$ is the number of cells in the `bottleneck' layer, and $L$ is the reconstruction error per country.

\subsection{Principal Component Analysis}

Principal Component Analysis~\cite{abdi2010principal} is a prevalent method for data compression and feature extraction. It uses orthogonal transformation to extract the high-dimensional data into a sequence of uncorrelated features.

The algorithm is defined as follows. For a data matrix $X$, we extract the first component by
\begin{equation}
w_{(1)} = \arg \max \left\{ \frac{w^T X^T X w}{w^T w} \right\}.
\end{equation}
Then we subsequently extract the $k$-th component given the first $k - 1$ ones by a subtracted matrix
\begin{equation}
X_{(k)} = X - \sum_{i = 1}^{k - 1} X w_{(i)} w_{(i)}^T,
\end{equation}
and a similar maximization of Rayleigh quotient
\begin{equation}
w_{(k)} = \underset{\|w\| = 1}{\arg \max} \left\{ \frac{w^T X_{(k)}^T X_{(k)} w}{w^T w} \right\}.
\end{equation}
The procedure could also be explained by the truncation of the largest singular values in the SVD decomposition. We use PCA to achieve an orthogonal design in the extracted feature space, in preparation for the regression.

\subsection{Linear Regression}

Since infectious diseases follow an exponential increase pattern, we take the logarithm of the death rate and find the data is of a Gaussian distribution. We shall perform linear regression
\begin{equation}
z = B a,
\end{equation}
where $z$ is the normalized log death rate, $B$ is the data of the features derived by autoencoders and PCA, augmented by a column of $1$ as the interception, and $a$ is our target of the weight of each feature.

\subsection{Higher-Order Rectification}

After the derivation of our regional model $z = B a$, where the weights are optimized by linear regression, we shall rectify the model by a higher-order perturbation,
\begin{equation}
z = B a + \epsilon M z,
\end{equation}
The matrix $M$ represents international flow of migration and trade, and $\epsilon$ is assumed small.

\section{Concrete Analysis}\label{sec:Concrete Analysis}

\subsection{Analysis of the Data After Completion}

\begin{figure}[tb]
\centering\includegraphics[width=\linewidth]{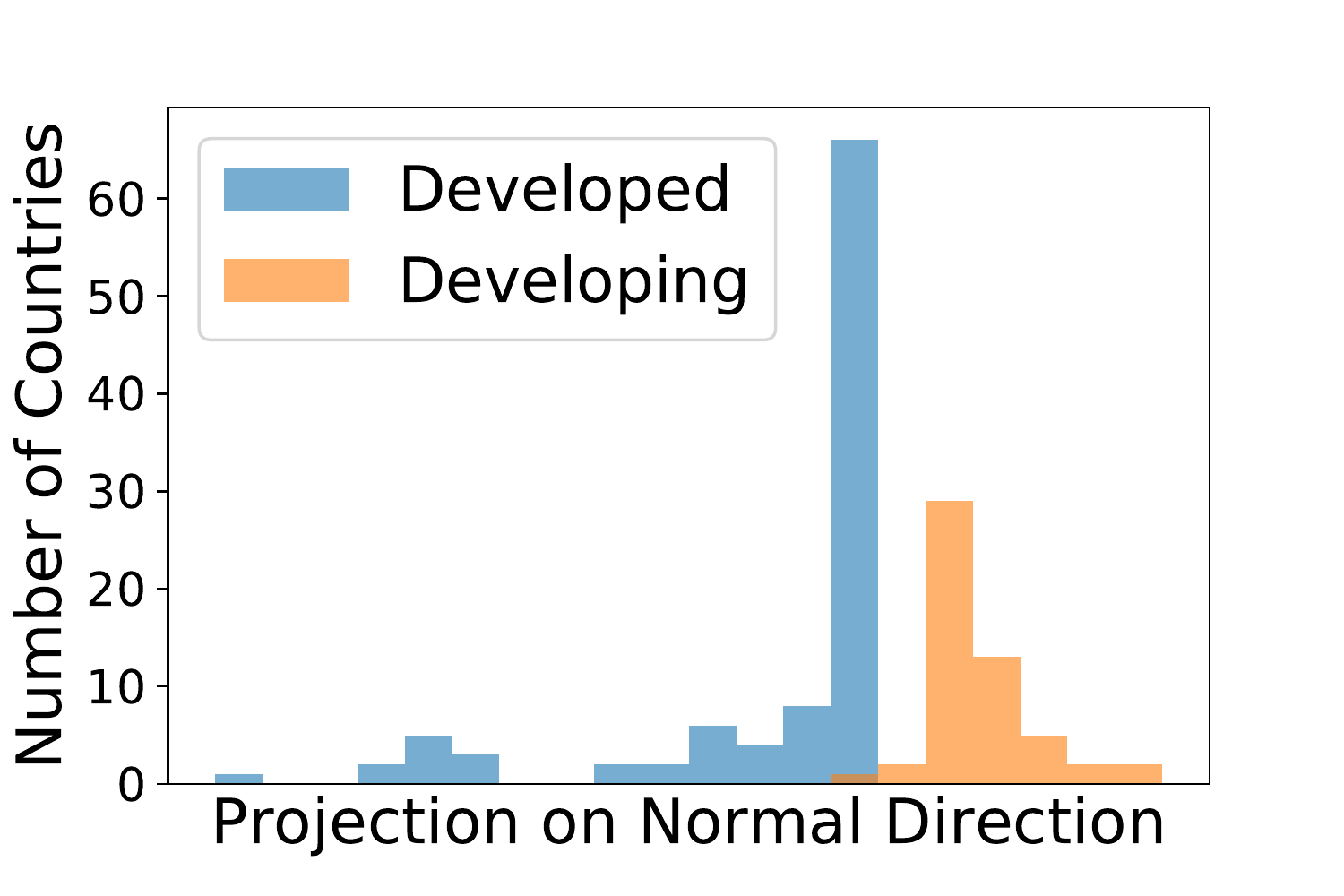}
\caption{The result of SVM for raw data}
\label{fig:SVM_raw}
\end{figure}

\begin{figure}[tb]
\centering\includegraphics[width=\linewidth]{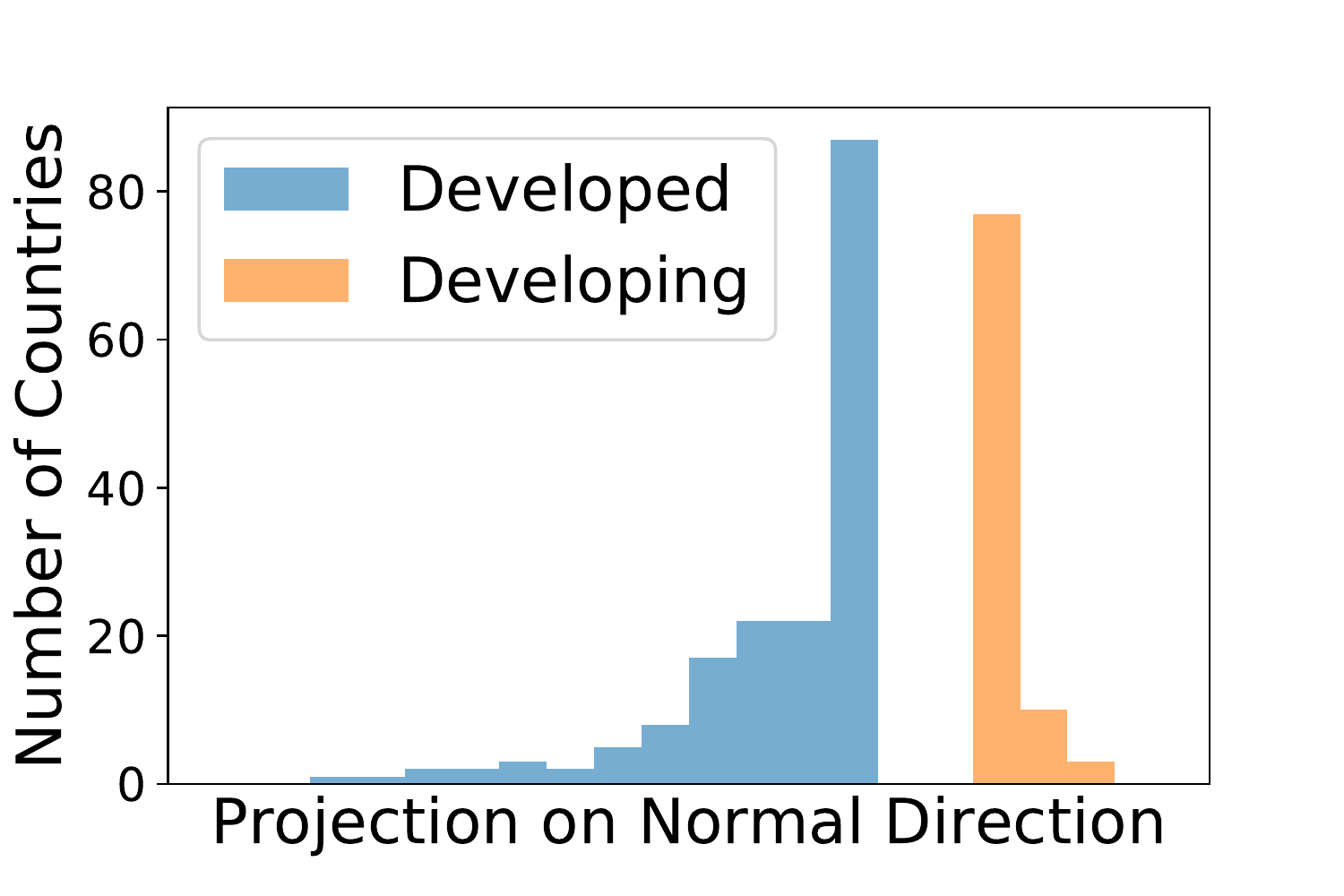}
\caption{The result of SVM for completed data}
\label{fig:SVM_completed}
\end{figure}

\begin{figure}[tb]
\centering\includegraphics[width=\linewidth]{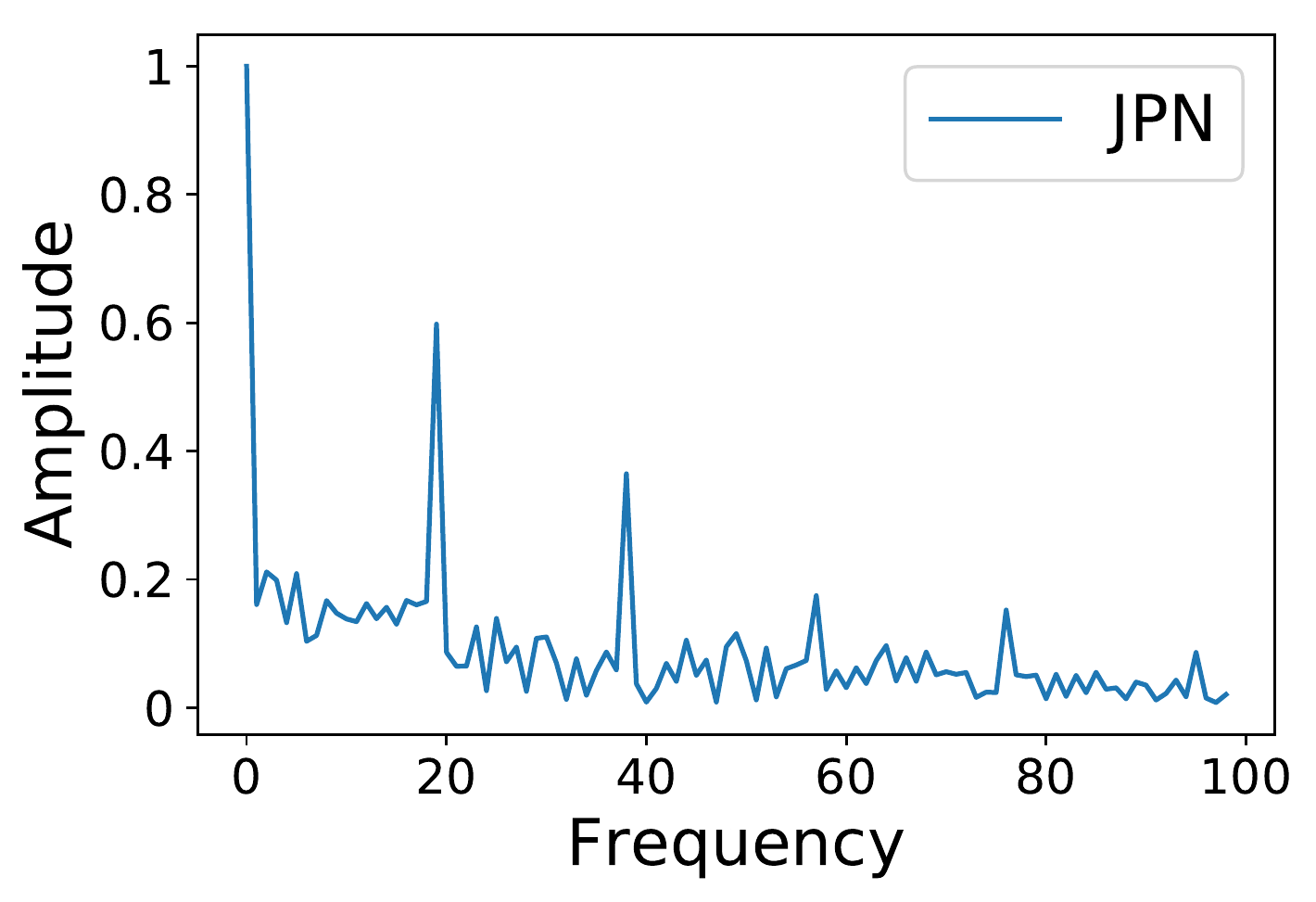}
\caption{Fourier series for influenza activity in Japan}
\label{fig:DFT_JPN}
\end{figure}

\begin{figure}[tb]
\centering\includegraphics[width=\linewidth]{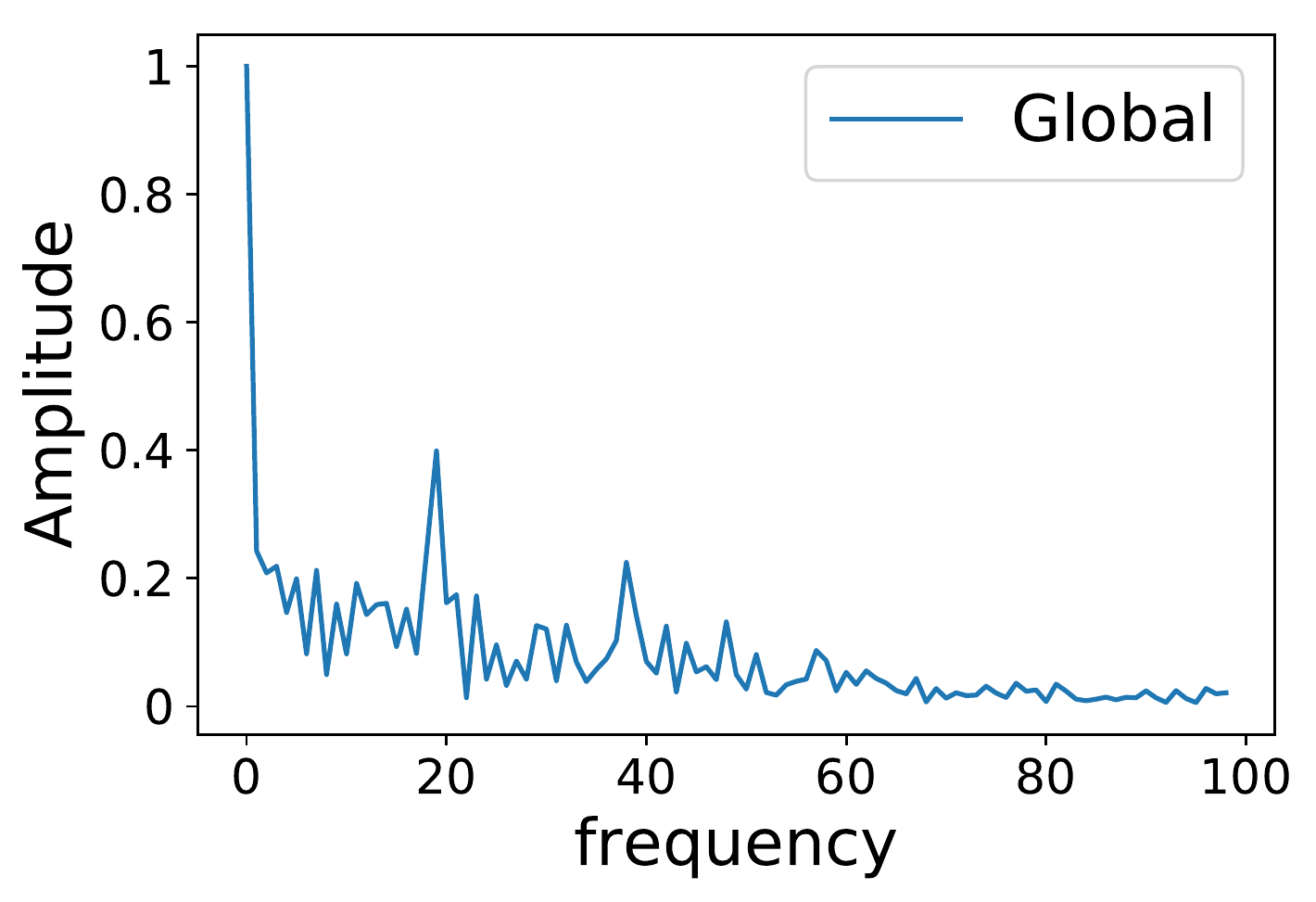}
\caption{Fourier series for influenza activity on earth}
\label{fig:DFT_GBL}
\end{figure}

We use the technique of matrix completion, for a merged version \texttt{world\_bank\_merge.csv} of the data \texttt{world\_bank} on World Bank~\cite{xixi}. The processed data \texttt{world\_bank\_impute.csv} is tested by a support vector machine. We would like to check the low-rank and sparse property of our data.

To be precise, we divide the countries into developed ones and developing ones, which is an essential feature indicative of the general condition of the countries. We use SVM for classification based on the completed features, and plot the results in Figure~\ref{fig:SVM_raw} and~\ref{fig:SVM_completed}. We find that the countries are successfully classified after completion, whereas the classification fails for the raw data, thus consolidating the effectiveness of our matrix completion.

Next, we use the discrete Fourier transform to analyze the periodic behavior in the dataset \texttt{influenza\_activity.csv}. The yearly periodic behavior can be easily spotted in Figure~\ref{fig:DFT_JPN} and~\ref{fig:DFT_GBL}, both regionally and globally, where $20$ in the horizontal axis stands for one year.

\subsection{Feature Extraction}

We use autoencoders to extract features from the originally $1170$-dimensional information \texttt{world\_bank\_impute.csv}. We use a four-layer autoencoder, with a number of selected features following a pattern of geometric series to capture the information better. Thus the number of features in each layer is $1170$, $356$, $108$, $33$, and finally $10$ on the final output layer.

\begin{figure}[htb]
\centering\includegraphics[width=\linewidth]{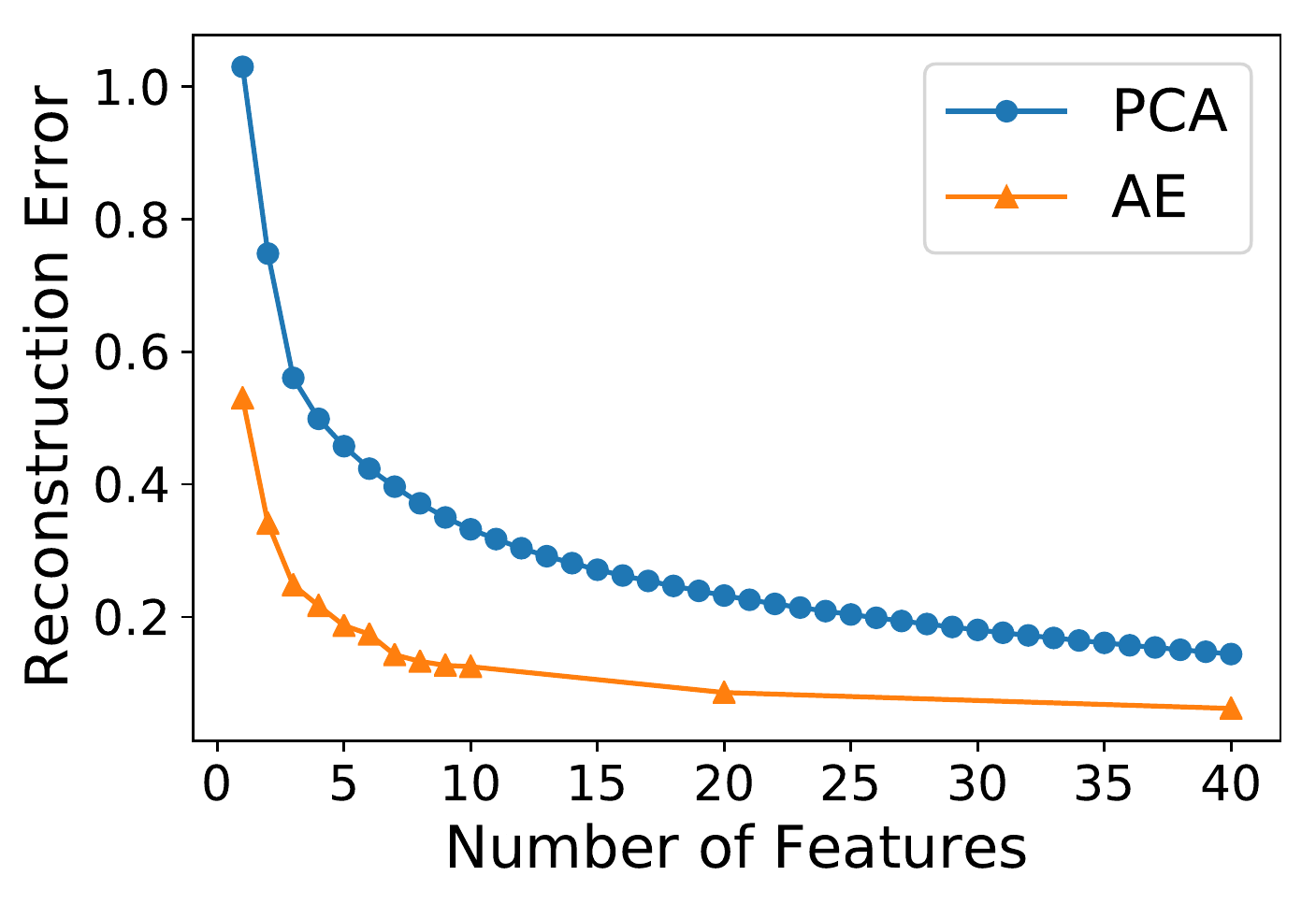}
\caption{Comparison of autoencoders with PCA}
\label{fig:ae_pca}
\end{figure}

The advantages of the autoencoder method over PCA lie in its nonlinearity and expressibility, which can be seen by comparing the reconstruction error of PCA and autoencoders in Figure~\ref{fig:ae_pca}. The error of autoencoders is far less than that of PCA given the same number of features. Thus autoencoders can indeed capture the nonlinearity of data features, which is beyond the abilities of PCA. However, taking the subsequent regression procedure into consideration, we shall invoke a PCA orthogonalization after the autoencoder.

Finally, we shall make some observations on our extracted features. The features are derived by a back-propagation of the results outputted by PCA, and they are linear combinations of the provided information. The specific data and corresponding information is stored in the zip file \texttt{Feature\_selection}, with each \texttt{.csv} file corresponding to one extracted feature, listed as per importance.

\subsection{Linear Regression and Higher-Order Rectification}

We take the influenza mortality data in \texttt{death\_rate\_ghe2016.csv} from WHO~\cite{ss}, and form the normalized log death rate $z$, stored in \texttt{z\_normal.txt}. We use a linear regression $z = B a$ to obtain the weights $a$, where $B$ is the feature matrix obtained by autoencoders and PCA.

As a higher-order rectification, we use the model
\begin{equation}
z = B a + \epsilon M z,
\end{equation}
where $\epsilon$ is assumed small by the intuition of perturbation. Each entry $M_{i j}$ represents the normalized mortality transferred from region $j$ to region $i$, which is composed of the flow amount multiplied by a region-dependent factor. The factor is further approximated by a second-order polynomial of $z$, inspired by the local rectangular basis in finite element method:
\begin{equation}
\begin{aligned}
M_{i j} &= (\alpha_0 + \alpha_1 z_i + \alpha_2 z_j + \alpha_3 z_i z_j) \, m_{i j} \\
&+ (\beta_0 + \beta_1 z_i + \beta_2 z_j + \beta_3 z_i z_j) \, t_{i j},
\end{aligned}
\end{equation}
where $\alpha_i$, $\beta_i$ are the coefficients to be optimized, and $m_{i j}$, $t_{i j}$ are normalized migration and trade amounts from region $j$ to region $i$.

The coefficients in the rectified version of our model can be jointly optimized with the linear regression, which essentially solves the approximated deconvolution problem. We store both of the outputs in $\texttt{out.txt}$.

As a final refinement of our model, we compute the $p$-value of each feature and shall reject the features with $p$-value more than $0.05$. The refined weights are listed in $\texttt{out\_selected.txt}$ and presented in Table~\ref{tab:weight}.

\begin{table}[htb]
\caption{Refined weights in significant features}
\label{tab:weight}
\centering
\begin{tabular}{cccc}
\toprule
Feature & $1$ & $2$ & $3$ \\
\midrule
Weight & $0.7709$ & $-0.2558$ & $-0.1083$ \\
\bottomrule
\end{tabular}
\end{table}

\begin{figure}[htb]
\centering\includegraphics[width=\linewidth]{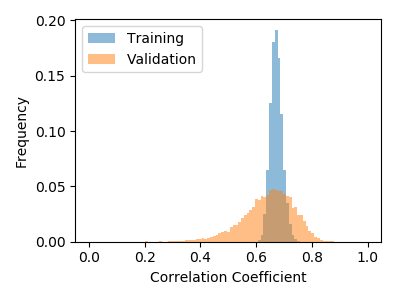}
\caption{Bootstrap test for linear regression model}
\label{fig:bootstrap}
\end{figure}

\begin{figure}[htb]
\centering\includegraphics[width=\linewidth]{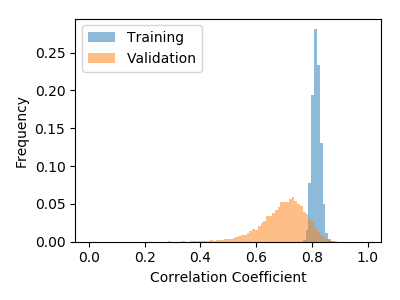}
\caption{Bootstrap test for model with rectification}
\label{fig:bootstrap_mt}
\end{figure}

A bootstrap cross-validation test is conducted to examine the behavior of our models. In Figure~\ref{fig:bootstrap} and~\ref{fig:bootstrap_mt}, we study the behavior of our model depending on whether the higher-order rectification is present. We can easily conclude that the modified model outperforms the original way by a considerable margin.

We now present some of our significant features and analyze their interpretation in real life scenarios, as in Table~\ref{table:info}.

\begin{table*}[htb]
\caption{Information contributing to the influenza mortality}
\centering
\begin{tabular}{cccc}
\toprule
Feature & 1 & 2 & 3 \\
\midrule
Bad & Other greenhouse gas emissions & Fertilizer consumption & Rural poverty gap \\
Medium & Electricity production & Employment to population ratio & Imports of goods and services \\
Good & Average precipitation in depth & Community health workers & Health care \\
\bottomrule
\end{tabular}
\label{table:info}
\end{table*}

\section{Policy Design}\label{sec:Policy}

\subsection{Global Policies}

Based on table~\ref{table:info}, we can design the following policies worldwide.
\begin{enumerate}
\item We can control the use of environmental-harmful products, such as fertilizers and electronic devices prone to greenhouse emission.
\item We should design policies to diminish the poverty gap, so as to provide the rural population with a better environment against influenza.
\item We can enhance our health welfare, spending more of government expenditure in health insurance and community health care to conquer the influenza virus.
\end{enumerate}

\subsection{Regional Policies}

In order to specifically design regional policies, we calculate the significant features for some typical countries, namely Nigeria, Japan and USA, as in Table~\ref{tab:weight_regional}.

\begin{table}[h]
\caption{Coefficients in Nigeria, Japan and USA for significant features}
\label{tab:weight_regional}
\centering
\begin{tabular}{cccc}
\toprule
Feature & $1$ & $2$ & $3$ \\
\midrule
Nigeria & $1.2139$ & $-1.0425$ & $-0.0322$ \\
Japan & $-0.9814$ & $0.0277$ & $1.0201$ \\
USA & $-0.3038$ & $0.3942$ & $-0.0076$ \\
\bottomrule
\end{tabular}
\end{table}

\begin{enumerate}
\item We analyze the data of Nigeria, where influenza is of the most mortality. We conclude that feature $1$ is of the most significance in the region, and we thus should impose restrictions on environmental-harmful activities such as the emission of greenhouse gases.
\item We carry out similar analysis on Japan, where influenza is of great significance in periodicity and the $86$-th least mortality. We conclude that feature $2$ should be enhanced in the region, and we thus should improve the number of community health workers. Japan is an isolated and highly populated country. Better community health care could lower the risk of spread of infectious diseases by intense contact.
\item Finally we study the cause for the spread of influenza in the United States, which has the $32$-nd lowest influenza mortality rate. We conclude that feature $3$ should be taken special notice in the region. Namely we should heed the poverty gap and provide better health care for the poor.
\end{enumerate}

The policies are indeed viable and can be propagated easily on internet. For example, we can put up slogans in marches and parades, to raise the general awareness of the public on environmental issues. The government should allocate more of its funding on community health care by adjusting its budgets. To address the problem of poverty gap, the government could increase the standard of minimal wages and impose relatively more taxes on the riches.

\section{Robustness and Regional Validation}\label{sec:Validation}

We introduce an additional column of information in \texttt{world\_bank\_merge.csv} with randomly generated data, to find that our results of feature extraction and weights regression remains similar, by a perturbation less than $5\%$. Similarly, an additional row of region is introduced in \texttt{world\_bank\_merge.csv} with randomly generated data, and the output remains a perturbation less than $5\%$ as well. Therefore, we can deduce that our model is of great robustness and stability.

Also, we compute the residual of our models in \texttt{residual.txt}, as regional validations of the models. We find that our model is of satisfactory accuracy, with moderate errors as model errors and algorithm errors.

\section{Assessment of Our Models}\label{Assessment}

\subsection{Strength of the Model}

\begin{enumerate}
\item The model consists of a linear regression of important features and a higher-order perturbation. Thus it enjoys the property of generalization and combines regional as well as international information.
\item Exterior datasets are introduced to help reconstruct missing information and features. Matrix completion ensures sparsity as well as low-rank property, matching the case in real-life scenarios. A high-dimensional data space ensures the comprehensiveness of our features, and model reduction methods are deployed to capture the crucial information.
\item DFT method is invoked to verify the periodic behavior of influenza.
\item Casual inference, cross-validation, and tests of robustness are executed for the sake of stability.
\item The model is transferable to model other types of infectious diseases due to its interpretability and its comprehensiveness of all potential contributing factors to influenza behavior.
\end{enumerate}

\subsection{Deficiencies of the Model}

\begin{enumerate}
\item We neglect the time evolution of influenza for lack of data.
\item A simplified version of the migration model is introduced, due to the complexity of modeling the dynamic system of a Markov process.
\item Policies are designed without taking into account the correlation between important features. Combined policies could be proposed as a future improvement.
\end{enumerate}

\section{Conclusion}\label{Conclusion}

We propose in this article a model of influenza spreading by a combination of massive feature engineering and international flow deconvolution. We detect a periodical behavior in influenza activity and use yearly normalized mortality data for regression. Features are extracted from the augmented dataset and weights are computed by higher-order rectification of graph deconvolution. We reach the conclusion that the spread of influenza is affected by its local environment and national economies. Policies are designed both regionally and globally, to mitigate the adverse effects of influenza.

Our model is of high interpretability. Bridging nonlinearity of the feature extraction into the linear regression model, supplemented by a higher-order graph deconvolution, can increase the robustness and consistency of the model. Furthermore, our methods of feature extraction and the main model can be easily transferred to analyze the spread of other infectious diseases.

\section*{Acknowledgement}
We thank Citadel and Correlation One for  all the dedicated efforts to hold such wonderful competition. This work is impossible without their help and support.

\newpage

\bibliographystyle{ieee}
\bibliography{ref}

\end{document}